\documentclass[letterpaper]{article}
\usepackage[preprint]{aaai2027}
\usepackage[hyphens]{url}
\usepackage{graphicx}
\usepackage{natbib}
\usepackage{caption}
\usepackage{booktabs}
\usepackage{amsmath}
\usepackage{amssymb}
\usepackage{listings}
\usepackage{courier}
\usepackage{textcomp}
\usepackage{placeins}
\usepackage[most]{tcolorbox}
\newcommand{\hostedsmall}{DeepSeek-V4.1-Flash}
\newcommand{\hostedlarge}{DeepSeek-V4-Pro}

\definecolor{promptframe}{HTML}{3F4A56}
\definecolor{promptback}{HTML}{F7F8FA}
\definecolor{promptvar}{HTML}{1F5F8B}
\newcommand{\ph}[1]{\textcolor{promptvar}{\textit{\{#1\}}}}
\newcommand{\role}[1]{\textcolor{promptframe}{\textsf{\bfseries\scriptsize\MakeUppercase{#1}}}}
\newtcblisting{promptbox}[2][]{%
  enhanced, breakable, listing only,
  colback=promptback, colframe=promptframe, colbacktitle=promptframe, coltitle=white,
  fonttitle=\small\bfseries, title={#2},
  boxrule=.5pt, arc=1.5pt, left=5pt, right=5pt, top=3pt, bottom=3pt,
  before skip=6pt, after skip=8pt,
  listing options={basicstyle=\ttfamily\footnotesize, breaklines=true,
    columns=fullflexible, keepspaces=true, breakindent=0pt,
    escapeinside={(*}{*)}, breakautoindent=false, breakatwhitespace=true, upquote=true, aboveskip=0pt, belowskip=0pt}, #1}

\title{Rare, Relevant, and Still Wrong: How Memory Fails LLM Group Decisions}
\author{
Anouar Imel\textsuperscript{\rm 1},
Imad Charradi\textsuperscript{\rm 2},
Abderrahman Sghir\textsuperscript{\rm 3},
Khalil El Khammar\textsuperscript{\rm 4}
}

\affiliations{
\textsuperscript{\rm 1}School of Information and Communication Engineering,
University of Electronic Science and Technology of China (UESTC)\\
\textsuperscript{\rm 2}Xidian University\\
\textsuperscript{\rm 3}Nanjing University of Aeronautics and Astronautics\\
\textsuperscript{\rm 4}School of Artificial Intelligence,
Nanjing University of Information Science and Technology\\
202524010105@std.uestc.edu.cn,
charradiimad@stu.xidian.edu.cn\\
abde44@nuaa.edu.cn,
khalil.elkhammar@nuist.edu.cn
}
\begin{document}
\maketitle
\begin{abstract}
When an LLM group's discussion is compressed into a short memory for a final decision maker, what should the memory keep? Two natural heuristics keep the rarest claims or the claims relevant to the decision criteria. On hidden-profile tasks both look attractive, because how widely a fact is held is confounded with whether it helps. We introduce CrossProfile, a diagnostic benchmark of 60 templates with automatically verified answer keys, in which the ownership of supporting, decoy, and irrelevant facts is crossed while their text is held fixed (480 tasks). Across Qwen2.5-7B, Llama-3.1-8B, and two DeepSeek models, both heuristics fail in predictable places. Selecting by relevance readmits decoys, because decoys are on topic: even a perfect relevance filter loses 23 to 31 points to rarest-first selection in the classic hidden-profile configuration. Rarest-first selection is no better than random selection on average and wins only where the rare facts happen to be the useful ones. In both cases accuracy follows the balance of evidence the memory keeps, not the policy that kept it. The decision model adds a third failure, a form of social proof: it gives extra weight to facts attributed to many agents. Removing the attribution tags, or reassigning them at random with line order fixed, raises accuracy in the classic configuration by 13 to 24 points. In natural discussions the evidence is lost before any memory acts: the complete transcript yields lower accuracy than no notes at all.
\end{abstract}

\section{Introduction}
An agent team may hold all the evidence needed for a decision and still leave its final reader with the wrong information. Some evidence is never stated. Other evidence is stated and then lost when a memory policy compresses the discussion. Even a faithful memory can emphasize claims that the reader weighs poorly. An end-task accuracy gain alone cannot tell these failure points apart.

\begin{figure}[!b]
\centering\includegraphics[width=.92\columnwidth]{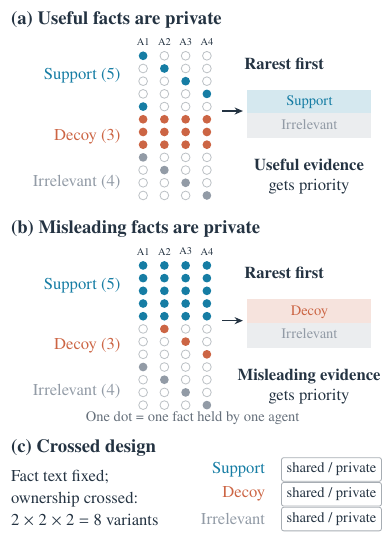}
\caption{\textbf{Rarity and decision usefulness.} Fact text stays fixed while the ownership of support and decoy facts changes; here irrelevant facts stay private. Filled dots mark the agents holding each fact, and the boxes on the right show what rarest-first selection prioritizes. (c) The three ownership factors are crossed. Decoys are true, criterion-relevant facts that favor a competing option. Role labels are never shown to the models.}
\label{fig:idea}
\end{figure}

Hidden-profile tasks make the distinction concrete: participants share some facts and individually hold other facts needed for a collective decision \citep{stasser1985pooling,gigone1993common,li2026hiddenbench}. A natural response to repeated shared information is to preserve what few participants said. We study this response through \emph{rarest-first selection} (R1), which ranks claims by the number of agents who stated them. The rule assumes that rare evidence is worth keeping.

Faithfulness alone does not settle which evidence to keep. In one procurement template, a supplier has passed every electrical safety inspection but took nine days to answer urgent repair callouts. Both statements concern the decision criteria. Keeping the favorable fact and dropping the unfavorable one can change the decision without introducing any false claim. What matters is the balance of the retained evidence, not only whether a summary is accurate.

Figure~\ref{fig:idea} shows why the rarity assumption needs testing. In a classic hidden profile, private evidence supports the correct answer and shared evidence favors a decoy. Prioritizing rare facts can then help by keeping support, by excluding decoys, or both. If private facts instead favor the decoy, the same rule can damage the decision. On HiddenBench the two explanations cannot be separated: a memory holding only the private facts reaches 91\% accuracy on Qwen, whereas one holding only the shared facts reaches 6\%, well below an empty memory. A large gain for rarest-first selection there therefore says little about rarity itself. The obvious alternative, keeping the facts relevant to the decision criteria, faces the opposite problem, because a decoy is relevant by construction. Testing either rule requires a design in which fact prevalence and decision usefulness vary independently.

We construct CrossProfile, a crossed benchmark in which support, decoy, and irrelevant facts are each either shared or private. Within a template, the fact sentences and their decision roles are fixed across all eight ownership variants, and the answer key follows from the fact roles and is verified automatically. Scripted discussions state every assigned fact, so memory selection can be studied against a known inventory. CrossProfile is a diagnostic benchmark: it isolates what a memory policy and a decision model do with a known set of evidence. Changing ownership also changes when a fact first appears, the holder tags attached to it, and its token cost, and we add a control for each of these.

Both heuristics fail in cells that the design predicts, and aggregate accuracy hides both failures. Relevance-first selection helps on average but loses 20 to 32 points in the classic hidden-profile cell, and a perfect relevance filter loses as much, because relevance does not distinguish support from decoys. Rarest-first selection is no better than random selection on average; its advantage over random selection changes by 28 to 41 points depending on who holds the decoys, on all four models. Across policies, accuracy rises steadily with the net evidence lead of the correct option among the kept facts, and once that lead is known the policy adds little. The decision model contributes a separate failure, a form of social proof: it gives extra weight to a fact attributed to many agents, although in our tasks shared facts were assigned, not independently observed. Removing the attribution tags, or reassigning them at random while keeping line order, raises its all-facts accuracy in the classic cell by 13 to 24 points. In natural discussions no memory policy beats another and the full transcript is less useful than an empty memory, so the evidence is lost before any memory acts.

Our contributions are (i) CrossProfile, a diagnostic benchmark that varies the ownership of support, decoy, and irrelevant facts while holding their content fixed; (ii) evidence on two open and two hosted models, with random-reference, fixed-item-count, and oracle-relevance controls, that relevance and rarity both fail in predictable ownership cells because neither tracks the balance of evidence; (iii) evidence that the decision model follows social proof in the form of holder tags, isolated from line order; and (iv) natural-discussion evidence that the main bottleneck lies upstream of memory.

\section{Related Work}
Hidden-profile research documents shared-information bias and the difficulty of pooling distributed evidence \citep{stasser1985pooling,gigone1993common,lu2012hidden}; in the common-knowledge effect, information held by more members carries more weight in the group judgment \citep{gigone1993common}, and our decision model shows a similar weighting from holder tags alone. LLM agents in opinion simulations can also violate basic probabilistic consistency, assigning probabilities to a claim and to its negation that do not add up to one \citep{imel2026truefalseaddone}. HiddenBench adapts this setting to LLM groups and examines forced disclosure, passive summarization, and structured communication \citep{li2026hiddenbench}. With human groups, an LLM facilitator increases how much information members share without improving the final decision \citep{alsobay2025table}, which parallels our finding that disclosure and use of evidence are separate bottlenecks. Our intervention holds fact content fixed while crossing the ownership of decision roles, and studies what survives a memory constraint applied after the discussion.

Multi-agent debate and agent frameworks study reasoning through interaction \citep{du2024debate,smit2024mad,wu2024autogen}, and collective biases in LLM groups can change non-linearly with group size \citep{flint2026groupsize}, which our fixed four-agent design does not vary. Consilience intervenes during hidden-profile communication with a calibrated controller that tracks evidence and redundancy \citep{babu2026consilience}. Communication interventions change what the group discloses; memory selection, which we study, determines which disclosed claims reach the final reader.

Memory architectures maintain information across interactions \citep{packer2023memgpt,park2023generative}, and prompt compression and selective context address limited capacity \citep{jiang2023llmlingua,li2023selective}. Benchmarks of conversational memory test long-horizon recall \citep{maharana2024locomo}, and recent multi-party benchmarks show that memory pipelines lose speaker and group structure \citep{yang2026groupmembench,hu2026evermembench}; CrossProfile instead asks which evidence a compressed memory should keep for a downstream decision. Interaction-aware selection argues that records can matter only in combination \citep{subrahmanyam2026beyond}, which is consistent with our finding that decisions follow the balance of the kept evidence rather than the merit of individual facts. Position effects further separate available evidence from evidence the model uses \citep{liu2024lost}; our tag controls hold line order fixed to separate attribution from position.

\section{Setting and Memory Policies}
\begin{figure*}[t]
\centering\includegraphics[width=\textwidth]{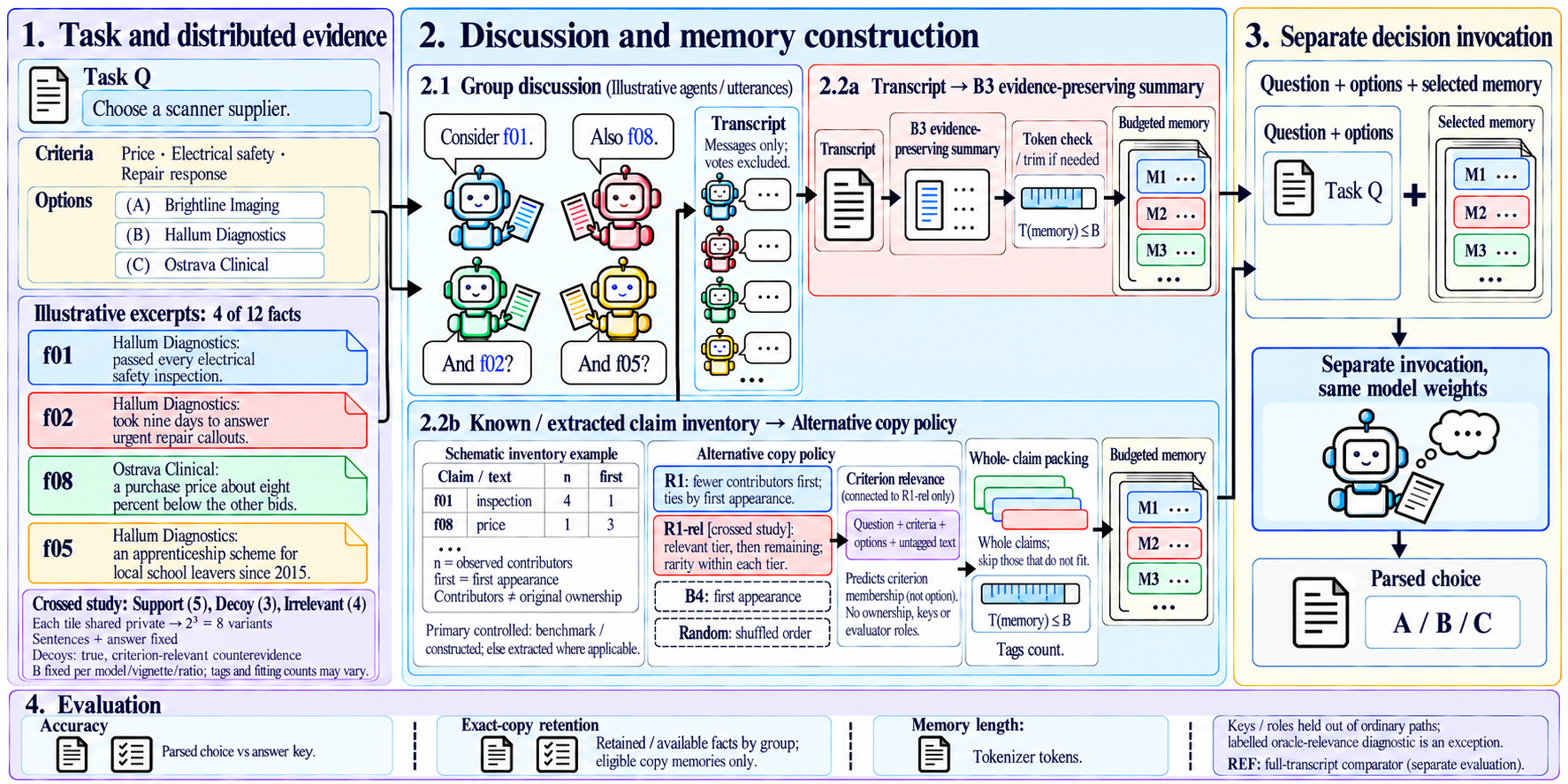}
\caption{\textbf{Pipeline.} Facts from template M01 (Table~\ref{tab:example}) are distributed among the agents; HiddenBench tasks have three or four agents and CrossProfile tasks have four. After the discussion, a memory policy compresses it under a token budget $B$, either by summarizing the transcript (B3) or by selecting whole claims from a claim inventory (R1, R1-rel, B4, Random). A separate call with the same model weights answers from the task and the memory alone.}
\label{fig:architecture}
\end{figure*}

\subsection{Discussion, Memory, and Decision}
A task specifies a description, candidate answers, and facts assigned to agents. A shared fact is held by every agent; a private fact is held by one. After the discussion, a memory policy constructs a memory for a separate decision call, which receives only the description, the options, and the memory. The decision call uses the same model weights as the discussion and sees the full transcript only in the REF control. Answer keys and role labels are withheld from all generation and from every memory policy except the oracle-relevance control.

In controlled HiddenBench discussions, each agent is scripted to state its assigned facts. Natural discussions consist of initial individual votes, two rounds of messages, and final votes; memories are built from the messages, not the votes. Copy policies select from a known claim inventory in the controlled condition and from an extracted inventory in the natural condition. In extracted inventories, a claim's contributors are the agents who stated it, which need not coincide with the agents who originally held the underlying fact. The summary policy B3 reads the transcript directly. Figure~\ref{fig:architecture} summarizes these paths.

\subsection{The CrossProfile Benchmark}
CrossProfile has 60 main templates, five in each of 12 decision domains. Each template has four agents, three options, and 12 facts: five support, three decoy, and four irrelevant. Support facts favor the correct option, including negative evidence about the decoy option. Decoy facts favor the competing option, including negative evidence about the correct option. Both concern the stated decision criteria; decoys are true statements whose partial implication is misleading. Irrelevant facts concern other attributes and may be positive or negative in tone. Each role group is independently shared or private, yielding $2^3$ ownership variants and 480 derived tasks. Ten pilot templates are excluded from all main estimates.

The answer key follows from the structured facts: each favorable fact adds one to its option's score, each unfavorable fact subtracts one, and irrelevant facts contribute zero. An automatic verifier checks every template for a unique best option over all facts, a decoy win when only decoy facts are counted, a correct-option win when only support facts are counted, and the absence of hint words; all 70 templates pass. With the fixed fact counts, the keyed option scores $+2$ over all facts and each competitor scores $0$. The task description asks for the best overall record on equally weighted criteria, but the decision prompt does not state the scoring rule. Accuracy therefore measures agreement with this constructed key.

The controlled transcript randomizes the speaking order and the order of each speaker's facts. Every shared fact appears in the first speaker's turn, although not necessarily before every private fact in that turn, so first-mention selection is sensitive to prevalence. Copied memory lines keep their contributor tags and follow the policy's selection order. Ownership thus changes how evidence is presented as well as its selection priority.

\subsection{Memory Policies and Budgets}
\textbf{R1} orders claims by increasing contributor count, breaking ties by first appearance. \textbf{B4} orders claims by first appearance alone. \textbf{Random} orders the same inventory uniformly at random. These copy policies add whole claims in this order until the budget is reached, skipping any claim that would exceed it.

\textbf{B3}, the evidence-preserving summary, asks the model to keep each distinct piece of evidence with its conditions, uncertainty, and disagreements. It receives a word limit derived from the budget, and longer outputs are truncated to the budget. \textbf{R1-rel} first asks the model which fact texts concern the stated criteria. This filter sees the description, options, and fact texts, but no holders, answers, or role labels. Claims predicted relevant form the first tier and the remaining claims the second, with R1 order inside each tier; second-tier claims remain eligible. Since decoys concern the criteria, they are relevant by this definition.

For model tokenizer $\tau_m$, the CrossProfile budget is
\begin{equation}\label{eq:budget}
B_{tm}(r)=\left\lfloor r\,|\tau_m(F_t^{\mathrm{all}})|\right\rfloor,
\qquad r\in\{.4,.6,.8\},
\end{equation}
where $F_t^{\mathrm{all}}$ is the text of every fact of template $t$ written with all four holder tags. The budget is therefore fixed across ownership variants and seeds. Selected lines carry their actual tags, so private facts cost fewer tokens and the same budget can hold different numbers of items. HiddenBench budgets are computed in the same way from the benchmark's fact set. The unbudgeted empty, all-facts, and full-transcript (REF) memories serve as controls. Appendix~\ref{app:measurement} additionally reports speaker-cycling selection (M1) on HiddenBench.

\subsection{Outcomes and Primary Contrasts}
Accuracy is averaged over seeds within each task or ownership cell, and then over tasks or templates. Invalid answers count as wrong. For CrossProfile template $t$, let $a_t(\pi\mid C)$ be the accuracy of policy $\pi$ averaged over seeds and over the variants satisfying ownership condition $C$. With $S,D,I$ denoting support, decoy, and irrelevant ownership, the primary contrasts at $r=.6$ are
\begin{equation}\label{eq:contrasts}
\begin{aligned}
h_{1t}&=[a_t(R1\mid D_s)-a_t(B4\mid D_s)]\\
&\quad-[a_t(R1\mid D_p)-a_t(B4\mid D_p)],\\
h_{2t}&=a_t(R1\mid I_s)-a_t(R1\mid I_p),\\
h_{3t}&=a_t(R1\text{-}rel\mid I_p)-a_t(R1\mid I_p),
\end{aligned}
\end{equation}
where subscripts $s$ and $p$ denote shared and private, and the remaining ownership factors are averaged within template. H1 tests whether the R1 advantage over B4 depends on decoy ownership, H2 tests a penalty when irrelevant facts are private, and H3 tests relevance-first priority when irrelevant facts are private.

For copy memories, retention is the fraction of each fact group that the memory contains, averaged within task and then over seeds. The retention gap is shared-fact retention minus private-fact retention, in percentage points. On HiddenBench, a fact counts as retained if the memory contains it verbatim after normalization; in CrossProfile, the selected facts are known exactly. Memory length is measured in tokens with the model's tokenizer.

\section{Experimental Setup}
We use Qwen2.5-7B-Instruct and Llama-3.1-8B-Instruct \citep{qwen2024qwen25,grattafiori2024llama3}, served locally with vLLM \citep{kwon2023vllm} in fp16 with an 8,192-token context and seeds 17, 42, and 91. Decisions, summaries, and claim extraction use temperature zero. Natural discussion messages use temperature 0.7, top-$p$ 0.9, and at most 256 output tokens; decisions allow 400 output tokens. HiddenBench contributes 65 tasks, 58 with four agents and 7 with three. We also evaluate two hosted models through the DeepSeek API at temperature zero with thinking disabled: \hostedsmall{} on all 60 CrossProfile templates and on natural HiddenBench discussions, and \hostedlarge{} on 30 templates. Appendices~\ref{app:protocol} and~\ref{app:reproduction} give all prompts and further implementation details.

\begin{figure*}[t]
\centering\includegraphics[width=\textwidth]{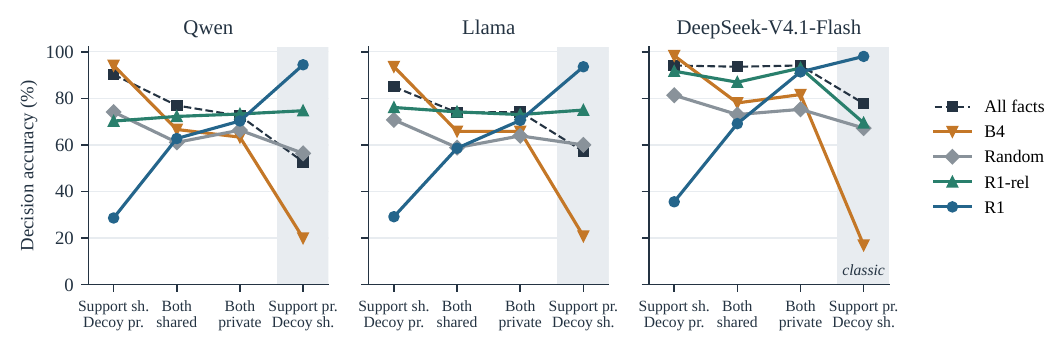}
\caption{\textbf{Who wins depends on who holds what.} CrossProfile accuracy at $r=.6$ by support and decoy ownership (each point averages the two irrelevant-fact assignments, 60 templates, three seeds). R1 and B4 are mirror images; R1-rel varies little across cells but falls well below R1 in the classic hidden-profile column (shaded: private support, shared decoys), where even the all-facts memory (dashed) drops. Per-cell intervals for Qwen and Llama are in Appendix~\ref{app:primary}.}
\label{fig:cells}
\end{figure*}

Both the HiddenBench and the CrossProfile experiments followed analysis plans specified before the main runs. In CrossProfile, $r=.6$ is the primary ratio and H1--H3 are the primary tests; other ratios, the reference-memory, oracle-relevance, item-count, and holder-tag controls, the hosted-model runs, and the prompt ablations are exploratory. A headroom check on the 65 HiddenBench tasks compares all-facts with empty memory, using a 4,096-token context. Qwen gains $+36.4$ points [25.1, 47.2], clearing the pre-specified 15-point threshold, whereas Llama's $+23.6$ [11.3, 35.9] does not clear it at the lower bound. Llama results on the original HiddenBench study are therefore treated as exploratory.

\paragraph{Inference.} We report 95\% percentile bootstrap intervals from 10,000 resamples. HiddenBench contrasts average seeds and ratios within task before resampling the 65 tasks; CrossProfile contrasts resample the 60 templates, keeping each template's variants and seeds together. For a paired difference, the two-sided $p$ value is $\hat p=\min\{1,2\min[\Pr^*(\bar d^*\leq0),\Pr^*(\bar d^*\geq0)]\}$, reported as $p<.001$ when no resample crosses zero. Holm correction \citep{holm1979simple} is applied to H1--H3 within each model at the primary ratio; intervals are not adjusted. The shared-irrelevant counterpart of H3 and all post-hoc controls lie outside this family. Appendix~\ref{app:measurement} lists the HiddenBench test families.

\begin{table}[t]
\centering
\small
\begin{tabular}{lrrrrr}
\toprule
\textbf{Policy} & \textbf{Acc.} & \textbf{Private} & \textbf{Shared} & \textbf{Gap} & \textbf{Tokens} \\
\midrule
\multicolumn{6}{l}{\textit{Qwen2.5-7B-Instruct}} \\
B3 & 51.3 & -- & -- & -- & 108 \\
B4 & 22.1 & 26.3 & 85.8 & 59.5 & 136 \\
R1 & 88.2 & 98.3 & 12.0 & $-86.3$ & 139 \\
Random & 42.1 & 52.2 & 58.4 & 6.2 & 136 \\
\midrule
\multicolumn{6}{l}{\textit{Llama-3.1-8B-Instruct}} \\
B3 & 59.0 & -- & -- & -- & 126 \\
B4 & 25.6 & 26.3 & 85.8 & 59.5 & 135 \\
R1 & 87.7 & 98.3 & 11.5 & $-86.8$ & 137 \\
Random & 43.1 & 53.9 & 56.7 & 2.8 & 133 \\
\bottomrule
\end{tabular}
\caption{Controlled HiddenBench at $r=.6$ (65 tasks, three seeds): decision accuracy (\%), exact-copy retention of private and shared facts (\%), retention gap (shared minus private, points), and median memory length (tokens).}
\label{tab:measurement}
\end{table}

\begin{table}[t]
\centering
\small
\begin{tabular}{lrr}
\toprule
\textbf{Memory} & \textbf{Qwen} & \textbf{Llama} \\
\midrule
Evidence summary (B3) & 72.0 & 58.7 \\
First mention (B4) & 61.0 & 61.5 \\
Random & 64.5 & 63.4 \\
Rarest first (R1) & 64.0 & 63.0 \\
Relevance first (R1-rel) & 72.6 & 74.6 \\
\midrule
All facts (unbudgeted) & 73.1 & 72.6 \\
Empty & 42.2 & 30.6 \\
\bottomrule
\end{tabular}
\caption{CrossProfile accuracy (\%) at $r=.6$, averaged over 60 templates, eight ownership variants, and three seeds.}
\label{tab:accuracy}
\end{table}

\section{Results}
\subsection{HiddenBench Confounds Rarity with Usefulness}
On controlled HiddenBench, R1 is far ahead of the other memories (Table~\ref{tab:measurement}). At $r=.6$ it reaches 88.2\% on Qwen and 87.7\% on Llama, against 51.3\% and 59.0\% for the evidence summary B3; averaged over budgets, R1 minus B3 is $+29.2$ points [22.9, 35.4] and $+26.7$ [20.2, 33.2], and rebuilding R1 at each summary's realized length leaves $+25.6$ and $+25.8$ (Appendix Table~\ref{tab:boundary}). Table~\ref{tab:measurement} also shows what R1 selects: nearly all private facts and few shared ones. Two reference memories show why this pays off. A memory of only the private facts reaches 91.3\% on Qwen and 90.3\% on Llama, above the all-facts memory by $+21.5$ [13.3, 30.3] and $+28.7$ [17.9, 39.0] points, whereas a memory of only the shared facts reaches 5.6\% and 4.6\%, far below an empty memory in the same run (35.4\% and 36.4\%). In HiddenBench the shared facts point to a wrong option, so dropping them is what helps, and no task has rare facts that mislead. CrossProfile adds such tasks.

\begin{figure*}[t]
\centering\includegraphics[width=\textwidth]{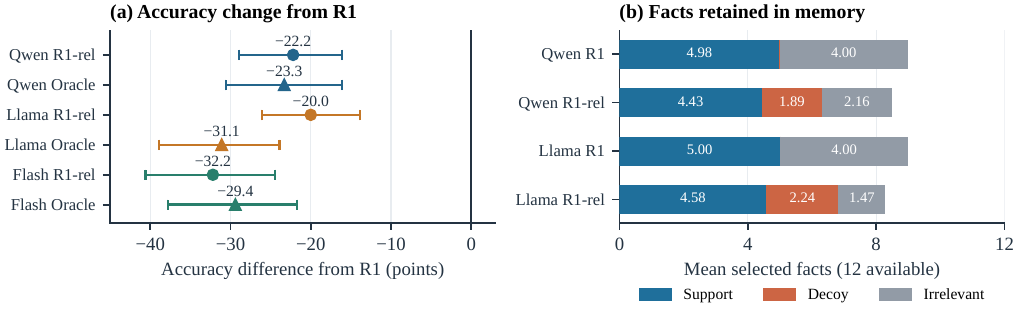}
\caption{\textbf{Criterion relevance is insufficient in the classic hidden-profile cell} (private support, shared decoys, private irrelevant facts, $r=.6$). (a) Accuracy differences from R1 for R1-rel and for oracle relevance on Qwen, Llama, and \hostedsmall{} (Flash), with 95\% template-bootstrap intervals. (b) Mean number of selected facts by role on Qwen and Llama: relevance-first priority admits shared decoys.}
\label{fig:counterexample}
\end{figure*}

\subsection{Rarity Is a Bet on Who Holds the Decoys}
Averaged over all ownership variants of CrossProfile, R1 is no better than Random: 64.0\% against 64.5\% on Qwen and 63.0\% against 63.4\% on Llama (Table~\ref{tab:accuracy}). This average hides a reversal across cells. When support is private and decoys are shared, R1 reaches 90 to 97\% while first-mention selection (B4) falls to 11 to 31\%; when support is shared and decoys are private, the order reverses, with B4 at 92 to 96\% and R1 at 18 to 40\% (Figure~\ref{fig:cells}).

The pre-specified H1 measures this dependence of the R1 advantage over B4 on decoy ownership: $+64.6$ points [60.3, 69.0] for Qwen and $+62.6$ [56.8, 68.2] for Llama, both Holm $p<.001$ (Appendix Figure~\ref{fig:primary}). Because every shared claim appears in the first turn, B4 keeps shared facts first and is close to a mirror image of R1, so part of H1 is built into the reference. Against Random, the interaction is $+40.7$ [34.6, 47.1] and $+34.2$ [27.8, 41.0]; when both policies retain exactly seven facts, H1 is $+66.8$ [61.9, 71.7] and $+72.6$ [67.2, 78.1]. The hosted models show the same pattern. On \hostedsmall{}, H1 is $+62.6$ [56.7, 68.6] and the interaction against Random is $+28.3$ [22.1, 34.7]; on a 30-template run of \hostedlarge{} the latter is $+30.6$ [22.5, 38.6]. On that common subset, R1 minus Random averaged over variants is $-1.7$, $+1.7$, $-1.3$, and $-4.7$ points on Qwen, Llama, \hostedsmall{}, and \hostedlarge{}.

Private irrelevant facts cost R1 little. H2 is $+5.0$ points [1.2, 8.9] for Qwen (Holm $p=.011$) and $+4.9$ [$-0.3$, 10.0] for Llama ($p=.060$). H2 varies with the budget, and H1 shrinks at $r=.8$, when the memory can hold most facts (Appendix~\ref{app:primary}).

\begin{figure}[t]
\centering\includegraphics[width=\columnwidth]{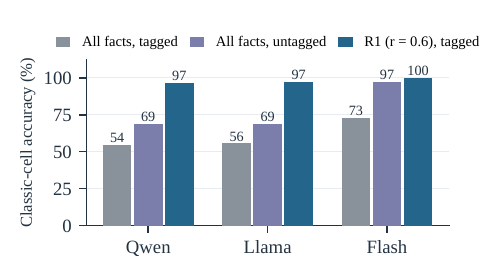}
\caption{\textbf{Holder tags mislead the decision model.} Accuracy in the classic cell (private support, shared decoys, private irrelevant facts) for the all-facts memory with and without holder tags, and for R1 at $r=.6$ (60 templates, three seeds; Flash: \hostedsmall{}).}
\label{fig:tags}
\end{figure}

\subsection{Relevant Is Not Useful}
H3 supports relevance-first priority on average: $+8.5$ points [3.5, 13.6] for Qwen (Holm $p=.002$), $+15.6$ [11.1, 19.9] for Llama ($p<.001$), and $+16.7$ [12.8, 20.7] on \hostedsmall{}, and R1-rel has the highest mean among budgeted policies (Table~\ref{tab:accuracy}). We expected no benefit when irrelevant facts are shared, but the gains are $+8.7$ [4.6, 12.8] and $+7.6$ [4.2, 11.2]: when all facts are shared, rarity cannot separate irrelevant facts from criterion-related ones, and the relevance tier does.

The average combines gains with a large loss. In the classic cell (private support, shared decoys, private irrelevant facts), R1 reaches 96.7\% [93.3, 99.4] on Qwen and 97.2\% [93.9, 99.4] on Llama, and R1-rel changes accuracy by $-22.2$ points [$-28.9$, $-16.1$] and $-20.0$ [$-26.1$, $-13.9$] (Figure~\ref{fig:counterexample}). The relevance tier admits about two shared decoys while retaining slightly less support; because shared decoys carry four holder tags, R1-rel also keeps fewer facts in this cell (8.3 to 8.5 against 9.0 for R1).

The failure persists when relevance is known exactly. With evaluator labels that mark both support and decoy facts as relevant, oracle relevance minus R1 is $-23.3$ [$-30.6$, $-16.1$] for Qwen and $-31.1$ [$-38.9$, $-23.9$] for Llama in this cell, so filter errors cannot explain the loss. An oracle that marked only support facts would no longer select by relevance; it would select by the answer key. The same holds on \hostedsmall{}, whose filter is nearly perfect (relevant-class precision 99.4\% and recall 98.9\%, against recall 66.0\% on Qwen and 85.6\% on Llama), yet in the same cell R1-rel minus R1 is $-32.2$ [$-40.6$, $-24.4$] and oracle relevance minus R1 is $-29.4$ [$-37.8$, $-21.7$]. Relevance separates the decision criteria from other attributes, whereas usefulness depends on the balance of evidence between competing options, which a relevance filter does not see.

The kept evidence explains these results across policies. Pooling the four copy policies, accuracy rises monotonically with the net lead of the correct option among the kept facts, counted as in the answer key: from 13.5\% and 10.3\% at a lead of $-2$ to 76.0\% and 75.6\% at $+2$ and 96.2\% and 96.1\% at $+4$ or more on Qwen and Llama. In a template random-intercept logistic model with this lead and the tone of the kept irrelevant facts, the policy terms (R1-rel, B4, and Random against R1) are no longer distinguishable from zero. Wrong answers are mostly the decoy: in the classic cell, B4 picks it in 71.1\% and 63.3\% of decisions. The decision model therefore follows the keyed evidence closely, and memory policies differ mainly in which balance of evidence they hand it.

\subsection{The Decision Model Follows Social Proof}
The all-facts memory is not a neutral reference. It contains every fact and gives the correct option a two-point lead in every variant, so a reader that ignored ownership would be equally accurate across cells. Instead, accuracy is 21.0 points [16.5, 25.6] higher on Qwen and 13.8 [9.7, 17.9] higher on Llama when support facts are shared rather than private. In the classic cell the all-facts memory reaches only 54.4\% and 55.6\%, about 40 points below the R1 memory built from the same facts under a 60\% budget (Figure~\ref{fig:tags}).

Each memory line begins with the tags of the agents who stated it, and much of this preference follows the tags (Figure~\ref{fig:tags}). Without tags, the shared-support advantage of the all-facts memory falls from $+21.0$ to $+7.2$ points on Qwen and from $+13.8$ to $+7.6$ on Llama. Removing them raises classic-cell accuracy by $+14.4$ points [7.8, 21.1] on Qwen, $+13.3$ [6.7, 20.0] on Llama, and $+24.4$ [17.2, 32.2] on \hostedsmall{}, where it goes from 72.8\% to 97.2\%. The tags, not the line order, carry this effect: both conditions keep the transcript-derived order, and reassigning the same tag sets to facts at random, again with order fixed, also raises classic-cell accuracy, by $+12.8$ points on both open models. The effect is not simply an error. Where support is shared and decoys are private, removing the tags lowers Qwen's accuracy by $-10.6$ points [$-16.7$, $-4.4$]: the model treats a fact stated by many agents as more credible, which helps when the widely held facts are the useful ones and misleads when they are not. In CrossProfile the shared facts were assigned to every agent rather than observed independently, so the tags carry no extra evidence. On a common 30-template subset the tagged gap is $+17.8$ on Qwen, $+12.5$ on Llama, $+6.1$ on \hostedsmall{}, and $+7.2$ on \hostedlarge{}. This attribution effect is an analogue of the common-knowledge effect that arises after the discussion rather than during it, and it accounts for part of R1's advantage in the classic cell: R1 drops the shared facts that the decision model overweights.

\subsection{Natural Discussions Lose the Evidence Upstream}
In natural discussions no memory policy is detectably better than another; R1 minus B3 is $-1.4$ points [$-5.0$, 2.4] on Qwen and $-3.1$ [$-9.2$, 2.7] on Llama (Appendix Table~\ref{tab:boundary}). The reason is visible before any memory is built (Table~\ref{tab:naturalbase}). The full transcript (REF) scores below an empty memory on all three models, and the group's own final vote is correct in only 24 to 32\% of discussions, although the same models reach 62 to 90\% when given every fact. In \hostedsmall{} discussions, a string-matching heuristic finds private facts stated in full in only 1.7\% [0.6, 2.9] of cases. Facts that were never stated cannot be recovered by any memory policy.

\begin{table}[t]
\centering
\small
\begin{tabular}{lrrrrr}
\toprule
\textbf{Model} & \textbf{Vote} & \textbf{REF} & \textbf{Empty} & \textbf{All} & \textbf{Discl.} \\
\midrule
Qwen & 23.6 & 22.1 & 33.8 & 70.3 & 0.0$^{\dagger}$ \\
Llama & 31.8 & 33.8 & 36.9 & 61.5 & 2.4$^{\dagger}$ \\
Flash & 24.6 & 25.6 & 35.9 & 89.7 & 1.7 \\
\bottomrule
\end{tabular}
\caption{Natural HiddenBench discussions (65 tasks, 195 discussions per model). Vote: accuracy of the group's majority vote after two rounds. REF, Empty, All: decision accuracy from the full transcript, an empty memory, and all facts (\%). Discl.: private facts stated in full in some message (\%, string-matching heuristic). $^{\dagger}$Disclosure for Qwen and Llama from one seed (65 discussions). Flash: \hostedsmall{}.}
\label{tab:naturalbase}
\end{table}

A prompt change helps one model only. Asking agents to state their evidence before recommending raises Llama's REF accuracy by $+30.8$ points [21.5, 39.5], and by $+25.1$ [15.9, 34.4] in a paired ablation with fresh seeds, with no detectable change for Qwen (Appendix~\ref{app:natural}). When the discussion carries more evidence, the memory policy matters again: under the evidence-first prompt, Llama's R1 memory reaches 57.1\% against 22.9\% for B4 and 53.2\% for B3, whereas in the original discussions the three were within three points of each other (29.2\%, 29.4\%, 32.3\%; means over budgets, Appendix~\ref{app:natural}).

\section{Conclusion}
Neither rarity nor relevance is a safe default for the memory of an LLM group. In CrossProfile, rarity helps only in the ownership cells where rare facts support the answer, relevance readmits on-topic decoys even with a perfect filter, and the decision model follows social proof, giving extra weight to facts that many agents are tagged as stating. What drives the decision is the balance of evidence between options that the memory leaves, and how the reader weighs its attribution. Memory designs for agent groups should therefore aim at that balance rather than at prevalence or topic, and should not present how many agents stated a fact as evidence for it when the sources are not independent. Evaluations should vary who holds which evidence, and should check whether that evidence reached the discussion at all.

\label{bodyend}
\FloatBarrier

\clearpage
\onecolumn
\appendix
\section*{Appendix A: Supplementary Results}
\setcounter{section}{1}
This appendix gives benchmark and policy details and all prompts (A.1), the full CrossProfile results (A.2), attribution controls (A.3), natural-discussion diagnostics and prompt ablations (A.4), measurement and statistical details (A.5), the hosted-model runs (A.6), and implementation details (A.7).

\subsection{Tasks, Policies, and Prompts}
\label{app:protocol}

\subsubsection*{Benchmark Construction and Example}
HiddenBench contributes 65 tasks: 58 have four agents and seven have three; 59 have three options and six have four. Its 503 facts comprise 250 shared and 253 private facts. Sixty facts contain several sentences and are kept as one unit. Controlled discussions disclose every assigned fact; natural discussions follow the benchmark's two-round exchange with initial and final votes.

CrossProfile has ten pilot and 60 main templates. The 60 main templates cover 12 decision domains with five templates each: hospital equipment purchase (M01--M05), school lunch vendor (M06--M10), city bus route (M11--M15), warehouse location (M16--M20), software hosting vendor (M21--M25), community clinic site (M26--M30), job candidate for a lab (M31--M35), conference venue (M36--M40), bridge repair contractor (M41--M45), farm irrigation system (M46--M50), library renovation (M51--M55), and emergency shelter site (M56--M60). Each main template has four agents, three options, and 12 facts. The support group contains three facts in favor of the correct option and two against the decoy option; the decoy group contains two facts in favor of the decoy and one against the correct option. The four irrelevant facts have two positive and two negative tones. The role structure is specified before the sentences are written, and facts receive neutral IDs in random order. Relevance refers to the stated criteria, not to whether a fact favors the answer.

Template M01 asks a hospital procurement panel to choose a scanner supplier, weighting purchase price, electrical safety, and repair response equally. The options are Brightline Imaging, Hallum Diagnostics, and Ostrava Clinical. Table~\ref{tab:example} lists its facts with their evaluator roles. The eight variants change the ownership of the role groups while keeping every sentence fixed.
\begin{table}[!htbp]
\centering\small
\begin{tabular}{llp{5.25in}}
\toprule
\textbf{ID} & \textbf{Role} & \textbf{Fact text} \\
\midrule
f01 & decoy & Hallum Diagnostics has passed every electrical safety inspection on its scanners. \\
f02 & support & Hallum Diagnostics took nine days to answer urgent repair callouts last year. \\
f03 & support & Ostrava Clinical passed every leakage test on its scanners during the last audit. \\
f04 & support & Ostrava Clinical engineers reach any ward within four hours for repairs. \\
f05 & irrelevant & Hallum Diagnostics has run an apprenticeship scheme for local school leavers since 2015. \\
f06 & irrelevant & Ostrava Clinical's headquarters was flooded in a burst pipe incident in 2019. \\
f07 & decoy & Ostrava Clinical failed a leakage test on one scanner batch during its previous contract. \\
f08 & support & Ostrava Clinical quoted a purchase price about eight percent below the other bids. \\
f09 & irrelevant & Brightline Imaging prints its manuals in large type for older nurses. \\
f10 & decoy & Hallum Diagnostics engineers reached ninety percent of repair callouts within two hours last year. \\
f11 & support & Hallum Diagnostics raised its invoice totals by eleven percent after the first year. \\
f12 & irrelevant & Ostrava Clinical's logo was criticised by a local design magazine last spring. \\
\bottomrule
\end{tabular}
\caption{All facts of template M01 (hospital scanner procurement; correct option: Ostrava Clinical; decoy: Hallum Diagnostics). Roles are evaluator annotations and are never given to the models.}
\label{tab:example}
\end{table}

\subsubsection*{Policy Details}
The claim inventory records each claim's text, its set of contributors, and its first-appearance rank. Each memory line is written as \texttt{[A1,A2] claim text}, and the memory keeps the selection order. R1 sorts by contributor count and then by first appearance; B4 uses first appearance; Random uses a random order. M1 cycles through speakers, taking each speaker's earliest unselected claim that fits, and then fills the remaining budget in first-appearance order. B3 summarizes the transcript instead of copying claims.

For R1-rel, the filter is queried once per template, seed, and model, and its prediction is shared by all eight ownership variants. Statements are presented in random order. If no valid list of statement numbers is obtained after one retry, all facts are treated as relevant, which reduces R1-rel to R1. The predicted-relevant and remaining tiers are each sorted in R1 order.

Copy policies add whole claims and skip claims that would exceed the budget. CrossProfile budgets price every fact with all four holder tags in template order, while selected lines carry their true tags, so the budget is the same across variants but the number of items that fit is not. Summaries are asked for at most $\max(1,\lfloor B/1.3\rfloor)$ words, and longer outputs are truncated to the budget at a sentence boundary.

\subsubsection*{Prompts}
The prompts below are used verbatim. Text in \ph{braces} is filled in for each call (task description, displayed options, transcript, memory, or numbered statements); no evaluator labels are ever inserted. Line breaks between filled-in components are shown schematically.

\begin{promptbox}{Decision call (all memory policies)}
(*\role{System}*)
You are a careful decision maker. Use only the information you are given.

(*\role{User}*)
(*\ph{task description}*)
(*\ph{options}*)
Notes from the group discussion. A tag such as [A1,A2] at the start of a line lists the participants who stated it:
(*\ph{memory}*)
Reply in exactly this format and nothing else:
Reasoning: <at most five sentences that weigh the options against the information above>
Decision: <one option copied exactly from the options above>

(*\role{Retry after an unparseable reply}*)
Your reply did not follow the format. Reply again with exactly two parts: 'Reasoning: ' followed by at most five sentences, then 'Decision: ' followed by one option copied exactly from the options list.
\end{promptbox}
A reply is valid if exactly one listed option can be parsed from it. Invalid replies are scored as wrong.

\begin{promptbox}{Evidence-preserving summary (B3)}
(*\role{System}*)
You write memories of group discussions. Use only the information in the transcript.

(*\role{User}*)
(*\ph{task description}*)
(*\ph{options}*)
Below is the transcript of a group discussion about this decision. Each message starts with the participant who wrote it, and messages are separated by a blank line.
(*\ph{transcript}*)
Write a memory of this discussion in at most (*\ph{W}*) words for someone who must make the decision later and cannot see the transcript. Keep every distinct piece of evidence that was mentioned, together with its conditions and qualifiers, any uncertainty, and any disagreement between participants. State each piece of evidence once. Do not add a recommendation of your own. Write plain text without headings. Output only the memory.
\end{promptbox}
\ph{W} is the word limit derived from the token budget.

\begin{promptbox}{Criterion-relevance filter (R1-rel)}
(*\role{System}*)
You sort statements by topic. Use only the information you are given.

(*\role{User}*)
(*\ph{task description}*)
(*\ph{options}*)
Below is a numbered list of statements collected in a group discussion about this decision.
(*\ph{numbered statements}*)
The description above names the decision criteria. Go through the statements one by one. A statement concerns a criterion if it tells us how an option performs on one of the stated criteria; a statement about anything else does not. Collect the numbers of the statements that concern a stated criterion and reply with one flat JSON list of those numbers, for example [2, 5, 7]. It must be a single list of integers, not an object and not grouped by criterion. Reply [] if there are none. Do not add any other text.

(*\role{Retry after an invalid reply}*)
Your reply was not one flat JSON list of statement numbers. Reply again with only a single JSON list such as [1, 4, 6], using only numbers from the list above, not grouped by criterion.
\end{promptbox}
The filter receives no contributor tags, roles, or answer.

\begin{promptbox}{Natural discussion (HiddenBench protocol)}
(*\role{System}*)
(*\ph{description}*)

You have received the following information. Notice that the order of these facts is randomly shuffled; the order does not indicate importance or relationship. Please reason carefully:
(*\ph{information}*)

Keep your response concise, just one or two sentences. (*\ph{extra}*)

(*\role{Initial vote}*)
Please decide and provide your rationale in the following JSON format:
{
    "vote": <A string, (*\ph{possible answers}*)>,
    "rationale": <A string, representing your rationale>
}

(*\role{First speaker, first round}*)
You are the first to speak.

(*\role{Other turns}*)
Previous messages from other people:
(*\ph{messages}*)
It is your turn to speak. (*\ph{extra}*)

(*\role{Final vote}*)
Previous messages from other people:
(*\ph{group discussion}*)
(*\ph{the initial-vote request above, verbatim}*)
\end{promptbox}
\ph{information} holds the agent's assigned facts and \ph{extra} is the benchmark's optional instruction field. After the initial votes, agents exchange two rounds of messages. In later rounds each agent sees the other agents' latest replies, including replies already updated in the current round; the final vote follows the last round. Memories are built from the messages, not from the votes. The prompt-ablation suffixes are given in Appendix~\ref{app:natural}.

In natural discussions, the claim inventory is built from the messages in two steps: every message is split into factual statements, and candidate pairs of statements with high word overlap are merged when the model judges them to state the same fact.

\begin{promptbox}{Claim extraction (natural inventories)}
(*\role{System}*)
You copy factual statements out of a message. Output JSON only.

(*\role{User}*)
This message was written in a group discussion:

<message>
(*\ph{message}*)
</message>

List every factual statement in the message.
Rules:
1. Copy each statement exactly as written. Do not paraphrase or join text from different places.
2. A statement is one fact together with its conditions and qualifiers. If a sentence only adds detail to the sentence before it, keep both sentences together as one statement. Example: "The bakery opens at 7. It also sells coffee." is one statement.
3. Leave out opinions, votes, recommendations and questions. If a sentence mixes a fact with an opinion, copy only the part that states the fact, exactly as written.
4. Keep the order of the message.
Output a JSON array of strings and nothing else, for example: ["first statement", "second statement"]

(*\role{Retry after an invalid reply}*)
Your reply was not a JSON array of strings copied from the message. Reply again with only the JSON array.
\end{promptbox}

\begin{promptbox}[unbreakable]{Duplicate check (natural inventories)}
(*\role{System}*)
You compare two statements from a group discussion. Answer with one word: yes or no.

(*\role{User}*)
Statement 1: (*\ph{first}*)
Statement 2: (*\ph{second}*)

Do these two statements state the same fact, with the same conditions and qualifiers? If one adds, drops or changes a detail, a number, a condition or a negation, the answer is no.
Answer yes or no.
\end{promptbox}

\FloatBarrier
\subsection{Full CrossProfile Results}
\label{app:primary}
Table~\ref{tab:allhyp} reports the three primary contrasts and the shared-irrelevant counterpart of H3 at every budget ratio. At $r=.6$, the Holm-adjusted $p$ values for H1, H2, and H3 are $<.001$, .011, and .002 for Qwen and $<.001$, .060, and $<.001$ for Llama.
\begin{table}[!htbp]
\centering
\small
\begin{tabular}{lrrrr}
\toprule
\textbf{Model / ratio} & \textbf{H1} & \textbf{H2} & \textbf{H3} & \textbf{H3, shared irrelevant} \\
\midrule
Qwen / .4 & $+65.3$ [60.1, 70.1] & $+13.1$ [9.7, 16.5] & $+20.8$ [16.0, 25.8] & $+8.1$ [3.9, 12.2] \\
Qwen / .6 & $+64.6$ [60.3, 69.0] & $+5.0$ [1.2, 8.9] & $+8.5$ [3.5, 13.6] & $+8.7$ [4.6, 12.8] \\
Qwen / .8 & $+35.3$ [30.3, 40.4] & $+2.4$ [$-1.1$, 5.7] & $+2.6$ [$-1.7$, 6.9] & $+2.9$ [$-0.6$, 6.4] \\
Llama / .4 & $+67.4$ [62.6, 72.1] & $+6.4$ [2.4, 10.4] & $+15.3$ [11.2, 19.4] & $+12.5$ [9.0, 16.0] \\
Llama / .6 & $+62.6$ [56.8, 68.2] & $+4.9$ [$-0.3$, 10.0] & $+15.6$ [11.1, 19.9] & $+7.6$ [4.2, 11.2] \\
Llama / .8 & $+29.4$ [24.7, 34.2] & $-1.5$ [$-5.3$, 2.4] & $+3.8$ [0.3, 7.4] & $+3.2$ [0.1, 6.4] \\
\bottomrule
\end{tabular}
\caption{CrossProfile contrasts (points) with 95\% template-bootstrap intervals. Only H1--H3 at $r=.6$ belong to the primary Holm family.}
\label{tab:allhyp}
\end{table}

\begin{figure}[!htbp]
\centering\includegraphics[width=\textwidth]{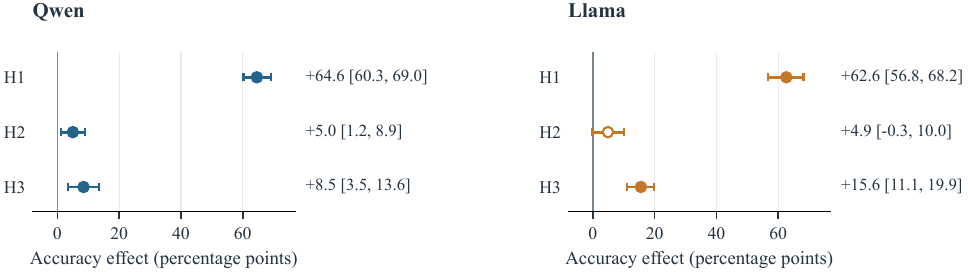}
\caption{\textbf{Primary contrasts at $r=.6$.} H1: dependence of the R1 advantage over B4 on decoy ownership. H2: R1 accuracy with shared minus private irrelevant facts. H3: R1-rel minus R1 with private irrelevant facts. Points and whiskers show estimates and 95\% template-bootstrap intervals. Filled markers are significant after Holm correction over H1--H3 within each model; Llama H2 is not.}
\label{fig:primary}
\end{figure}

\begin{table}[!htbp]
\centering
\small
\begin{tabular}{lrrrrrr}
\toprule
& \multicolumn{3}{c}{\textbf{Qwen}} & \multicolumn{3}{c}{\textbf{Llama}} \\
\cmidrule(lr){2-4}\cmidrule(lr){5-7}
\textbf{Policy} & $r=.4$ & $r=.6$ & $r=.8$ & $r=.4$ & $r=.6$ & $r=.8$ \\
\midrule
B3 & 68.9 & 72.0 & 72.6 & 52.8 & 58.7 & 60.6 \\
B4 & 51.5 & 61.0 & 68.0 & 50.5 & 61.5 & 70.9 \\
Random & 57.8 & 64.5 & 71.6 & 55.1 & 63.4 & 70.6 \\
R1 & 54.6 & 64.0 & 71.9 & 53.6 & 63.0 & 72.6 \\
R1-rel & 69.0 & 72.6 & 74.7 & 67.5 & 74.6 & 76.0 \\
\midrule
All facts & 73.1 & 73.1 & 73.1 & 72.6 & 72.6 & 72.6 \\
Empty & 42.2 & 42.2 & 42.2 & 30.6 & 30.6 & 30.6 \\
\bottomrule
\end{tabular}
\caption{CrossProfile accuracy (\%) at all budget ratios.}
\label{tab:budgetgrid}
\end{table}

\begin{figure}[!htbp]
\centering\includegraphics[width=\textwidth]{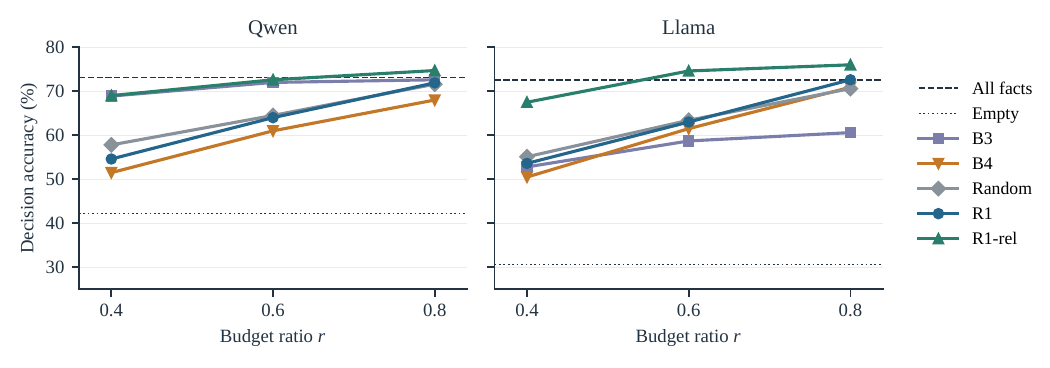}
\caption{CrossProfile accuracy across budget ratios (Table~\ref{tab:budgetgrid}). Points are means over templates, variants, and seeds; horizontal lines mark the unbudgeted all-facts and empty controls.}
\label{fig:budgets}
\end{figure}

\begin{figure}[!htbp]
\centering\includegraphics[width=.90\textwidth]{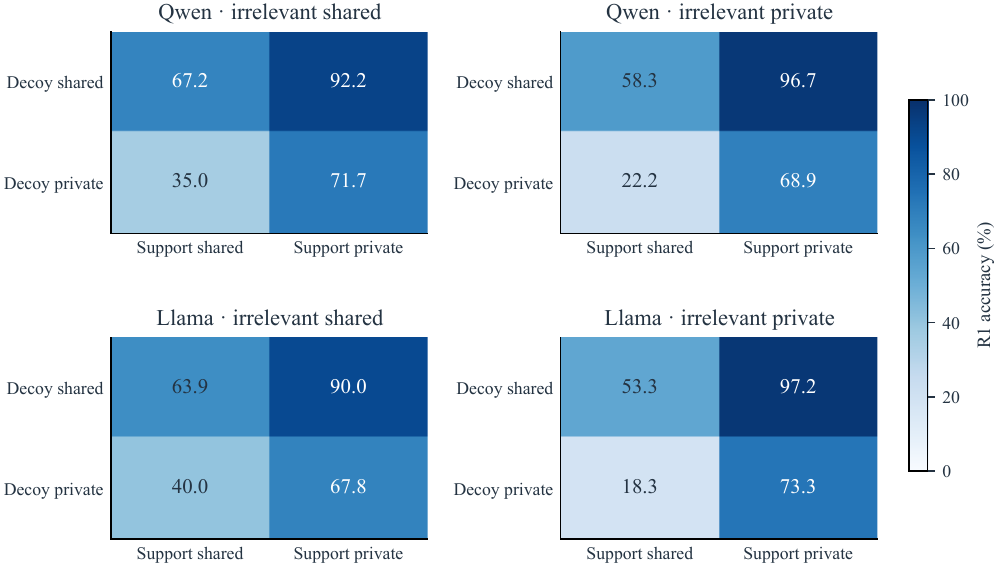}
\caption{R1 accuracy at $r=.6$ by support and decoy ownership, separately for each model and irrelevant-fact assignment. Every cell averages the same 60 templates and three seeds; intervals are in Tables~\ref{tab:cellsQwenselection} and~\ref{tab:cellsLlamaselection}.}
\label{fig:heatmap}
\end{figure}

Tables~\ref{tab:cellsQwenselection}--\ref{tab:cellsLlamacontrols} give accuracy in each ownership cell at $r=.6$ with 95\% intervals; Figure~\ref{fig:heatmap} visualizes the R1 column. Empty memory does not depend on ownership and scores 42.2\% [36.1, 48.3] for Qwen and 30.6\% [23.9, 37.8] for Llama in every cell.
\begin{table}[!htbp]
\centering
\small
\begin{tabular}{lrrrr}
\toprule
\textbf{S / D / I} & \textbf{R1} & \textbf{R1-rel} & \textbf{B4} & \textbf{Random} \\
\midrule
sh / sh / sh & 67.2 [60.0, 74.4] & 75.6 [68.9, 81.7] & 67.2 [60.0, 74.4] & 60.0 [52.2, 67.8] \\
sh / sh / pr & 58.3 [50.6, 66.1] & 68.9 [60.6, 76.7] & 66.1 [57.8, 74.4] & 62.2 [54.4, 69.4] \\
sh / pr / sh & 35.0 [28.3, 41.7] & 73.3 [66.1, 80.0] & 92.2 [88.3, 96.1] & 72.2 [64.4, 79.4] \\
sh / pr / pr & 22.2 [14.4, 30.6] & 67.2 [60.6, 73.9] & 96.1 [92.8, 98.9] & 76.1 [68.9, 83.3] \\
pr / sh / sh & 92.2 [87.8, 96.1] & 75.0 [68.3, 81.7] & 13.9 [8.3, 20.0] & 58.9 [51.7, 66.1] \\
pr / sh / pr & 96.7 [93.3, 99.4] & 74.4 [67.8, 80.6] & 26.1 [19.4, 33.3] & 53.9 [46.1, 61.7] \\
pr / pr / sh & 71.7 [64.4, 78.3] & 77.2 [71.1, 83.3] & 56.7 [48.3, 65.0] & 63.3 [55.6, 70.6] \\
pr / pr / pr & 68.9 [62.2, 75.6] & 69.4 [63.3, 75.6] & 70.0 [62.8, 76.7] & 69.4 [61.7, 76.7] \\
\bottomrule
\end{tabular}
\caption{Qwen, copy policies: accuracy (\%) by ownership cell at $r=.6$, with 95\% template-bootstrap intervals. S/D/I: ownership of support, decoy, and irrelevant facts (sh: shared, pr: private).}
\label{tab:cellsQwenselection}
\end{table}

\begin{table}[!htbp]
\centering
\small
\begin{tabular}{lrr}
\toprule
\textbf{S / D / I} & \textbf{B3} & \textbf{All facts} \\
\midrule
sh / sh / sh & 72.2 [65.0, 79.4] & 78.9 [71.7, 85.6] \\
sh / sh / pr & 81.7 [76.1, 87.2] & 75.0 [68.3, 81.7] \\
sh / pr / sh & 86.1 [79.4, 91.7] & 89.4 [83.9, 94.4] \\
sh / pr / pr & 91.1 [86.7, 95.0] & 91.1 [86.1, 95.6] \\
pr / sh / sh & 46.7 [40.0, 53.9] & 51.1 [42.2, 60.0] \\
pr / sh / pr & 56.1 [47.2, 65.0] & 54.4 [46.1, 62.8] \\
pr / pr / sh & 71.1 [64.4, 77.8] & 75.0 [67.8, 82.2] \\
pr / pr / pr & 71.1 [62.8, 78.3] & 70.0 [62.8, 76.7] \\
\bottomrule
\end{tabular}
\caption{Qwen, summary and all-facts memories: accuracy (\%) by ownership cell at $r=.6$, with 95\% template-bootstrap intervals. Notation as in Table~\ref{tab:cellsQwenselection}.}
\label{tab:cellsQwencontrols}
\end{table}

\begin{table}[!htbp]
\centering
\small
\begin{tabular}{lrrrr}
\toprule
\textbf{S / D / I} & \textbf{R1} & \textbf{R1-rel} & \textbf{B4} & \textbf{Random} \\
\midrule
sh / sh / sh & 63.9 [56.1, 71.7] & 72.8 [66.7, 78.9] & 62.8 [55.0, 70.6] & 58.3 [50.0, 66.1] \\
sh / sh / pr & 53.3 [45.0, 61.1] & 75.6 [69.4, 81.7] & 68.9 [61.7, 76.1] & 59.4 [52.8, 66.1] \\
sh / pr / sh & 40.0 [31.7, 48.9] & 73.9 [67.2, 80.6] & 91.7 [87.2, 95.6] & 69.4 [61.7, 77.2] \\
sh / pr / pr & 18.3 [12.2, 24.4] & 78.3 [71.7, 84.4] & 95.6 [92.2, 98.3] & 72.2 [65.0, 79.4] \\
pr / sh / sh & 90.0 [85.0, 94.4] & 72.8 [65.0, 80.0] & 10.6 [6.1, 15.0] & 57.8 [49.4, 65.6] \\
pr / sh / pr & 97.2 [93.9, 99.4] & 77.2 [70.0, 83.9] & 31.1 [23.9, 38.3] & 62.2 [54.4, 70.0] \\
pr / pr / sh & 67.8 [59.4, 75.6] & 72.8 [65.0, 80.6] & 58.3 [50.0, 66.7] & 65.6 [57.8, 73.3] \\
pr / pr / pr & 73.3 [65.6, 81.1] & 73.3 [66.1, 80.0] & 73.3 [65.6, 81.1] & 62.2 [53.3, 71.1] \\
\bottomrule
\end{tabular}
\caption{Llama, copy policies: accuracy (\%) by ownership cell at $r=.6$, with 95\% template-bootstrap intervals. Notation as in Table~\ref{tab:cellsQwenselection}.}
\label{tab:cellsLlamaselection}
\end{table}

\begin{table}[!htbp]
\centering
\small
\begin{tabular}{lrr}
\toprule
\textbf{S / D / I} & \textbf{B3} & \textbf{All facts} \\
\midrule
sh / sh / sh & 63.3 [55.6, 70.6] & 72.2 [64.4, 80.0] \\
sh / sh / pr & 65.0 [56.7, 73.3] & 75.6 [67.8, 83.3] \\
sh / pr / sh & 71.1 [64.4, 77.8] & 85.0 [77.8, 91.7] \\
sh / pr / pr & 71.1 [62.2, 79.4] & 85.0 [78.9, 90.6] \\
pr / sh / sh & 38.3 [31.1, 45.6] & 58.9 [50.0, 67.8] \\
pr / sh / pr & 50.6 [43.3, 58.3] & 55.6 [47.2, 63.3] \\
pr / pr / sh & 55.6 [47.8, 63.3] & 73.3 [65.6, 81.1] \\
pr / pr / pr & 54.4 [46.1, 62.2] & 75.0 [68.3, 81.1] \\
\bottomrule
\end{tabular}
\caption{Llama, summary and all-facts memories: accuracy (\%) by ownership cell at $r=.6$, with 95\% template-bootstrap intervals. Notation as in Table~\ref{tab:cellsQwenselection}.}
\label{tab:cellsLlamacontrols}
\end{table}

\FloatBarrier
\subsection{Attribution Controls}
\label{app:attribution}
Table~\ref{tab:robustness} collects the control analyses. Replacing B4 by Random as the reference tests whether H1 depends on first-mention ordering. The seven-item experiment fixes the number of selected facts at $k=\lfloor .6\times12\rfloor=7$ for both policies. Oracle relevance uses the evaluator roles to mark support and decoy facts as criterion-relevant. The tag experiment either removes the holder tags from all-facts memories or reassigns the same tag sets to facts at random; both keep the transcript-derived line order, so they separate attribution from position. Without tags the shared-support gap is $+7.2$ on Qwen and $+7.6$ on Llama; with shuffled tags it is $+14.6$ and $+6.4$.
\begin{table}[!htbp]
\centering
\small
\begin{tabular}{lrr}
\toprule
\textbf{Contrast} & \textbf{Qwen} & \textbf{Llama} \\
\midrule
H1 with Random as reference & $+40.7$ [34.6, 47.1] & $+34.2$ [27.8, 41.0] \\
H1 with seven retained items & $+66.8$ [61.9, 71.7] & $+72.6$ [67.2, 78.1] \\
Oracle relevance $-$ R1, classic cell & $-23.3$ [$-30.6$, $-16.1$] & $-31.1$ [$-38.9$, $-23.9$] \\
Tagged $-$ untagged support gap, all facts & $+13.8$ [8.1, 19.3] & $+6.1$ [1.0, 11.2] \\
Tagged $-$ shuffled-tag support gap, all facts & $+6.4$ [1.9, 11.1] & $+7.4$ [2.2, 12.6] \\
Untagged $-$ tagged, classic cell, all facts & $+14.4$ [7.8, 21.1] & $+13.3$ [6.7, 20.0] \\
Shuffled $-$ tagged, classic cell, all facts & $+12.8$ [3.9, 21.7] & $+12.8$ [5.6, 20.0] \\
\bottomrule
\end{tabular}
\caption{Attribution controls (points) with 95\% template-bootstrap intervals. Classic cell: private support, shared decoys, private irrelevant facts.}
\label{tab:robustness}
\end{table}

\paragraph{Selection and filter behavior.} In the classic cell, Qwen's R1 memories contain on average 4.98 support, 0.01 decoy, and 4.00 irrelevant facts, against 4.43, 1.89, and 2.16 under R1-rel. For Llama the corresponding values are 5.00, 0.00, and 4.00 against 4.58, 2.24, and 1.47. The relevance filter recalls 76.7\% of criterion-relevant facts that favor an option and 48.1\% of those that disfavor one on Qwen, and 94.6\% and 70.7\% on Llama.

\FloatBarrier
\subsection{Natural Discussions and Prompt Ablations}
\label{app:natural}
Five diagnostics each change one component relative to the original natural runs (N0): N1 labels observations with origin-neutral IDs; N2 removes the reminder of the agent's initial vote; N3 appends the evidence-first instruction of arm A1 below to the discussion prompt; N4 extracts claims at the sentence level; and N5 merges duplicate claims only when their wording is identical. N1--N3 regenerate the discussions, whereas N4 and N5 rebuild only the B4 and R1 memories from the original discussions. Qwen shows no detectable change under any diagnostic (Table~\ref{tab:diagQwen}). For Llama, N3 raises REF by $+30.8$ points [21.5, 39.5], R1 by $+27.9$ [20.2, 35.7], and B3 by $+20.9$ [14.2, 27.7] (Table~\ref{tab:diagLlama}).
\begin{table}[!htbp]
\centering
\small
\begin{tabular}{lrrrrr}
\toprule
\textbf{Policy} & \textbf{N1} & \textbf{N2} & \textbf{N3} & \textbf{N4} & \textbf{N5} \\
\midrule
B3 & $-1.9$ [$-7.7$, 4.1] & $-1.0$ [$-8.7$, 6.7] & $+2.1$ [$-5.3$, 9.6] & -- & -- \\
B4 & $-1.2$ [$-7.5$, 5.1] & $+1.4$ [$-5.3$, 8.2] & $-3.1$ [$-9.4$, 3.1] & $-0.2$ [$-3.2$, 2.9] & 0.0 [$-1.9$, 1.9] \\
R1 & $-2.9$ [$-8.5$, 2.7] & $-0.7$ [$-7.5$, 6.3] & $+1.4$ [$-5.3$, 7.9] & $+1.0$ [$-2.4$, 4.4] & $+0.3$ [$-1.9$, 2.4] \\
REF & $+0.5$ [$-5.6$, 6.7] & $+0.5$ [$-7.2$, 8.2] & $+6.2$ [$-1.5$, 13.8] & -- & -- \\
\bottomrule
\end{tabular}
\caption{Qwen: change in natural-discussion accuracy relative to the original runs (points), averaged over budget ratios, with 95\% task-bootstrap intervals. Dashes mark memories that N4 and N5 do not rebuild.}
\label{tab:diagQwen}
\end{table}

\begin{table}[!htbp]
\centering
\small
\begin{tabular}{lrrrrr}
\toprule
\textbf{Policy} & \textbf{N1} & \textbf{N2} & \textbf{N3} & \textbf{N4} & \textbf{N5} \\
\midrule
B3 & $+1.4$ [$-4.6$, 7.3] & $+5.1$ [$-1.2$, 11.5] & $+20.9$ [14.2, 27.7] & -- & -- \\
B4 & $+3.8$ [$-3.2$, 10.9] & $+6.5$ [$-0.2$, 13.2] & $-6.5$ [$-12.7$, $-0.2$] & $-2.7$ [$-8.2$, 2.4] & $-0.3$ [$-2.9$, 2.2] \\
R1 & $+5.0$ [$-1.0$, 11.1] & $+3.4$ [$-3.6$, 10.8] & $+27.9$ [20.2, 35.7] & $-2.6$ [$-7.9$, 2.7] & 0.0 [$-2.1$, 1.9] \\
REF & $-1.0$ [$-9.7$, 7.2] & $-1.0$ [$-8.7$, 6.7] & $+30.8$ [21.5, 39.5] & -- & -- \\
\bottomrule
\end{tabular}
\caption{Llama: change in natural-discussion accuracy relative to the original runs (points), averaged over budget ratios, with 95\% task-bootstrap intervals. Dashes mark memories that N4 and N5 do not rebuild.}
\label{tab:diagLlama}
\end{table}

\paragraph{Prompt ablation.} To separate the components of the evidence-first wording, we ran a paired ablation with fresh seeds 7, 23, and 58, in which every arm regenerates the discussions of all 65 tasks. The outcome is REF accuracy from the full transcript, so the ablation concerns the discussion rather than any compression policy. Arm A0 uses the original prompt; arms A1--A4 append the suffixes below to the first-round user prompt.

\begin{promptbox}[unbreakable]{Discussion-prompt suffixes for arms A1--A4}
(*\role{A1: Evidence first, no recommendation}*)
Before recommending an option, state the concrete observations in the information you received. Preserve their conditions, negations and numbers. Do not recommend an option in this turn.

(*\role{A2: Evidence first}*)
Before recommending an option, state the concrete observations in the information you received. Preserve their conditions, negations and numbers.

(*\role{A3: No recommendation}*)
Do not recommend an option in this turn.

(*\role{A4: Neutral length request}*)
Please take your time and write a full, careful reply that covers what you know.
\end{promptbox}

\begin{table}[!htbp]
\centering
\small
\begin{tabular}{lrrrr}
\toprule
& \multicolumn{2}{c}{\textbf{Qwen}} & \multicolumn{2}{c}{\textbf{Llama}} \\
\cmidrule(lr){2-3}\cmidrule(lr){4-5}
\textbf{Arm} & \textbf{Accuracy} & \textbf{$\Delta$ vs.\ A0} & \textbf{Accuracy} & \textbf{$\Delta$ vs.\ A0} \\
\midrule
A0: original & 30.3 [22.6, 37.9] & -- & 30.8 [23.1, 39.0] & -- \\
A1: evidence first, no recommendation & 28.2 [21.5, 35.4] & $-2.1$ [$-10.3$, 5.6] & 59.0 [50.8, 67.7] & $+28.2$ [17.9, 39.0] \\
A2: evidence first & 35.9 [28.2, 43.6] & $+5.6$ [$-1.0$, 12.3] & 55.9 [48.2, 63.6] & $+25.1$ [15.9, 34.4] \\
A3: no recommendation & 22.6 [15.9, 29.7] & $-7.7$ [$-15.9$, 0.0] & 40.5 [33.3, 47.7] & $+9.7$ [0.5, 18.5] \\
A4: neutral length request & 31.3 [23.6, 39.5] & $+1.0$ [$-4.1$, 6.7] & 38.5 [31.8, 45.6] & $+7.7$ [$-1.5$, 16.9] \\
\bottomrule
\end{tabular}
\caption{Discussion-prompt ablation: REF accuracy (\%) and paired differences from A0 (points), over 65 tasks and seeds 7, 23, and 58, with 95\% task-bootstrap intervals.}
\label{tab:natural}
\end{table}

\begin{figure}[!htbp]
\centering\includegraphics[width=\textwidth]{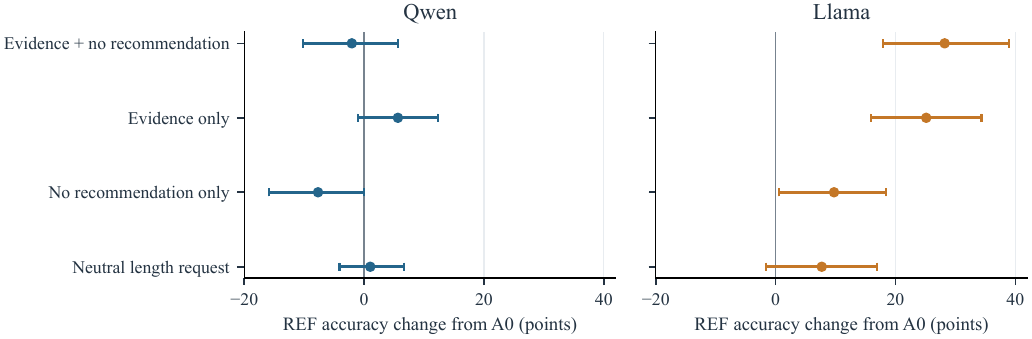}
\caption{Discussion-prompt ablation: change in REF accuracy relative to A0, with 95\% task-bootstrap intervals. Qwen does not reproduce Llama's large gain from evidence-first wording.}
\label{fig:natural}
\end{figure}

The difference between A1 and A2 on Llama is $+3.1$ points [$-6.2$, 12.3].

\FloatBarrier
\subsection{Measurement and Statistical Details}
\label{app:measurement}
For a fixed policy and budget, task accuracy averages the three seeds. HiddenBench contrasts additionally average the budget ratios within task. CrossProfile accuracy averages seeds within a template--variant pair and then the relevant variants within template. Bootstrap samples resample tasks or whole templates, never individual variants. Invalid answers are scored as wrong.

For paired differences $d_t$, we draw 10,000 task or template resamples and take the 2.5th and 97.5th percentiles of the resampled mean. The two-sided bootstrap $p$ value is
\begin{equation}
\hat p=\min\{1,2\min[\Pr^*(\bar d^*\leq0),\Pr^*(\bar d^*\geq0)]\},
\end{equation}
with $p=1$ when all differences are zero; values below .001 are reported as $p<.001$. Holm's step-down correction is applied within the pre-specified families. On HiddenBench, each family contains R1 $-$ B3, R1 $-$ B4, and M1 $-$ B4 for one model, discussion condition, and inventory type, with ratios averaged within task. Table~\ref{tab:frozenfull} lists all contrasts. ``Known'' inventories are the benchmark's fact lists used in controlled discussions; ``extracted'' inventories are built by the model from the transcript.
\begin{table}[!htbp]
\centering
\small
\begin{tabular}{llrr}
\toprule
\textbf{Model} & \textbf{Condition / contrast} & \textbf{Effect [95\% CI]} & \textbf{Holm $p$} \\
\midrule
Qwen & Headroom: all facts $-$ empty & $+36.4$ [25.1, 47.2] & -- \\
Qwen & Controlled, known: R1 $-$ B3 & $+29.2$ [22.9, 35.4] & $<.001$ \\
Qwen & Controlled, known: R1 $-$ B4 & $+55.4$ [49.4, 61.4] & $<.001$ \\
Qwen & Controlled, known: M1 $-$ B4 & $-4.6$ [$-8.2$, $-1.2$] & .008 \\
Qwen & Controlled, extracted: R1 $-$ B3 & $+25.8$ [18.8, 32.6] & $<.001$ \\
Qwen & Controlled, extracted: R1 $-$ B4 & $+53.2$ [47.5, 58.6] & $<.001$ \\
Qwen & Controlled, extracted: M1 $-$ B4 & $-2.9$ [$-6.3$, 0.5] & .094 \\
Qwen & Natural: R1 $-$ B3 & $-1.4$ [$-5.0$, 2.4] & 1.000 \\
Qwen & Natural: R1 $-$ B4 & $-0.5$ [$-2.6$, 1.5] & 1.000 \\
Qwen & Natural: M1 $-$ B4 & $+1.2$ [$-1.7$, 4.1] & 1.000 \\
\midrule
Llama & Headroom: all facts $-$ empty & $+23.6$ [11.3, 35.9] & -- \\
Llama & Controlled, known: R1 $-$ B3 & $+26.7$ [20.2, 33.2] & $<.001$ \\
Llama & Controlled, known: R1 $-$ B4 & $+52.5$ [46.3, 58.1] & $<.001$ \\
Llama & Controlled, known: M1 $-$ B4 & $-4.6$ [$-8.0$, $-1.4$] & .006 \\
Llama & Controlled, extracted: R1 $-$ B3 & $+18.1$ [11.5, 25.0] & $<.001$ \\
Llama & Controlled, extracted: R1 $-$ B4 & $+44.4$ [37.1, 51.5] & $<.001$ \\
Llama & Controlled, extracted: M1 $-$ B4 & $-2.2$ [$-7.0$, 2.6] & .376 \\
Llama & Natural: R1 $-$ B3 & $-3.1$ [$-9.2$, 2.7] & .888 \\
Llama & Natural: R1 $-$ B4 & $-0.2$ [$-3.2$, 2.9] & 1.000 \\
Llama & Natural: M1 $-$ B4 & $+0.3$ [$-4.1$, 4.8] & 1.000 \\
\bottomrule
\end{tabular}
\caption{All HiddenBench headroom and memory contrasts (points).}
\label{tab:frozenfull}
\end{table}

\begin{table}[!htbp]
\centering
\small
\begin{tabular}{lrr}
\toprule
\textbf{Comparison} & \textbf{Qwen} & \textbf{Llama} \\
\midrule
Controlled: R1 $-$ B3 & $+29.2$ [22.9, 35.4] & $+26.7$ [20.2, 33.2] \\
Controlled: R1 at B3 length $-$ B3 & $+25.6$ [17.6, 33.7] & $+25.8$ [19.1, 32.5] \\
Natural: R1 $-$ B3 & $-1.4$ [$-5.0$, 2.4] & $-3.1$ [$-9.2$, 2.7] \\
\bottomrule
\end{tabular}
\caption{HiddenBench accuracy differences (points) with 95\% task-bootstrap intervals over 65 tasks; seeds and the three budget ratios are averaged within task. The controlled R1 $-$ B3 contrast belongs to the pre-specified Holm family.}
\label{tab:boundary}
\end{table}

HiddenBench exact-copy retention checks whether each normalized fact is contained as a whole in the normalized memory. Facts are averaged within task and then over seeds and tasks. CrossProfile retention is computed from the selected facts and their roles. Table~\ref{tab:measurement} in the main text reports the HiddenBench retention values.

Table~\ref{tab:utilization} reports the mean ratio of realized memory tokens to the budget in CrossProfile; these are means, unlike the medians in Table~\ref{tab:measurement}. The R1-at-B3-length comparison uses each summary's actual length as the budget.
\begin{table}[!htbp]
\centering
\small
\begin{tabular}{lrr}
\toprule
\textbf{Model / policy} & \textbf{Tokens / budget (\%)} & \textbf{95\% CI} \\
\midrule
Qwen / R1 & 95.0 & [94.8, 95.3] \\
Qwen / R1-rel & 95.7 & [95.5, 95.8] \\
Qwen / B4 & 95.8 & [95.7, 96.0] \\
Qwen / Random & 95.6 & [95.5, 95.8] \\
Qwen / B3 & 79.3 & [77.8, 80.8] \\
\midrule
Llama / R1 & 95.1 & [94.9, 95.4] \\
Llama / R1-rel & 95.8 & [95.7, 96.0] \\
Llama / B4 & 95.8 & [95.7, 95.9] \\
Llama / Random & 95.5 & [95.3, 95.6] \\
Llama / B3 & 85.5 & [84.9, 86.0] \\
\bottomrule
\end{tabular}
\caption{Realized memory length relative to the budget in CrossProfile at $r=.6$, averaged within template, with 95\% template-bootstrap intervals.}
\label{tab:utilization}
\end{table}

\FloatBarrier
\subsection{Hosted Models and Fixed Budget}
\label{app:hosted}
Table~\ref{tab:hosted} reports the hosted-model contrasts. \hostedsmall{} H1--H3 are Holm-corrected within that run. The classic cell is the cell with private support, shared decoys, and private irrelevant facts, as in the main text. The \hostedlarge{} run covers templates M01--M30 and four memory conditions (R1, Random, all facts, empty); R1-rel was not run on \hostedlarge{}. Averaged over variants, R1 minus Random on this subset is close to zero for every model ($-1.7$, $+1.7$, $-1.3$, and $-4.7$ points for Qwen, Llama, \hostedsmall{}, and \hostedlarge{}), while its dependence on decoy ownership is large for all four. The natural-discussion run on \hostedsmall{} includes only REF, all-facts, and empty memories.

\begin{table}[!htbp]
\centering
\small
\begin{tabular}{llrr}
\toprule
\textbf{Run} & \textbf{Contrast} & \textbf{Effect [95\% CI]} & \textbf{Templates / tasks} \\
\midrule
\hostedsmall{}, CrossProfile & H1 & $+62.6$ [56.7, 68.6] & 60 \\
\hostedsmall{}, CrossProfile & H2 & $+11.5$ [7.5, 15.6] & 60 \\
\hostedsmall{}, CrossProfile & H3 & $+16.7$ [12.8, 20.7] & 60 \\
\hostedsmall{}, CrossProfile & H3, shared irrelevant & $+6.8$ [3.2, 10.4] & 60 \\
\hostedsmall{}, CrossProfile & H1 with Random as reference & $+28.3$ [22.1, 34.7] & 60 \\
\hostedsmall{}, CrossProfile & R1-rel $-$ R1, classic cell & $-32.2$ [$-40.6$, $-24.4$] & 60 \\
\hostedsmall{}, additional controls & Oracle relevance $-$ R1, all variants & $+11.7$ [8.2, 15.1] & 60 \\
\hostedsmall{}, additional controls & Oracle relevance $-$ R1, classic cell & $-29.4$ [$-37.8$, $-21.7$] & 60 \\
\hostedsmall{}, additional controls & Untagged $-$ tagged all facts, classic cell & $+24.4$ [17.2, 32.2] & 60 \\
\hostedsmall{}, additional controls & Tagged $-$ untagged support gap & $+12.5$ [8.3, 17.1] & 60 \\
\hostedsmall{}, natural & REF $-$ empty & $-10.3$ [$-19.5$, $-1.0$] & 65 \\
\hostedsmall{}, natural & Private facts stated in full (\%) & 1.7 [0.6, 2.9] & 65 \\
\hostedlarge{}, CrossProfile subset & H1 with Random as reference & $+30.6$ [22.5, 38.6] & 30 \\
\hostedlarge{}, CrossProfile subset & R1 $-$ Random, all variants & $-4.7$ [$-8.2$, $-1.2$] & 30 \\
\hostedlarge{}, CrossProfile subset & Tagged support gap, all facts & $+7.2$ [3.3, 11.4] & 30 \\
\bottomrule
\end{tabular}
\caption{Hosted-model contrasts (points) with 95\% bootstrap intervals.}
\label{tab:hosted}
\end{table}

\begin{table}[!htbp]
\centering
\small
\begin{tabular}{lrr}
\toprule
\textbf{Run} & \textbf{Valid} & \textbf{Invalid} \\
\midrule
\hostedsmall{}, CrossProfile & 8,466 & 174 \\
\hostedsmall{}, additional controls & 4,184 & 136 \\
\hostedsmall{}, natural & 564 & 21 \\
\hostedlarge{}, CrossProfile subset & 2,880 & 0 \\
\bottomrule
\end{tabular}
\caption{Hosted-model decision counts. As in the local runs, a decision is invalid only after one format retry, and invalid decisions are scored as wrong.}
\label{tab:counts}
\end{table}

\paragraph{Fixed budget.} To check that the controlled R1 advantage over B3 does not depend on defining budgets as a ratio of the fact set, we also ran both policies with a fixed 128-token budget on controlled HiddenBench. There, R1 and B3 reach 81.0\% and 52.3\% on Qwen and 82.1\% and 57.4\% on Llama. This experiment used a 4,096-token context rather than the 8,192-token context of the main runs.

\subsection{Implementation Details}
\label{app:reproduction}
Local models are served with vLLM on two NVIDIA T4 GPUs. \hostedsmall{} and \hostedlarge{} are accessed through the DeepSeek API (model identifiers \texttt{deepseek-flash} and \texttt{deepseek-v4-pro}). The order in which options are displayed rotates across seeds.

We will release the code, the templates, and all generated discussions, memories, and model outputs upon publication.

\FloatBarrier
\end{document}